\documentclass[10pt,twocolumn,letterpaper]{article}
\usepackage[pagenumbers]{cvpr}
\usepackage[T1]{fontenc}
\usepackage{amsmath,amssymb,amsthm}
\usepackage{graphicx,booktabs,multirow,tabularx}
\usepackage{xcolor}
\usepackage{microtype}
\usepackage{enumitem}
\usepackage{algorithm}
\usepackage{algpseudocode}
\definecolor{cvprblue}{rgb}{0.21,0.49,0.74}
\usepackage[pagebackref,breaklinks,colorlinks,allcolors=cvprblue]{hyperref}
\usepackage[capitalize,noabbrev]{cleveref}
\newcommand{\method}{GAAFormer}
\newcommand{\serial}{Serial-OUFormer}

\newcommand{\N}{\mathcal{N}}
\newcommand{\T}{\mathcal{T}}
\newcommand{\U}{\mathcal{U}}

\newcommand{\smallhead}[1]{\noindent\textbf{#1}\enspace}
\newtheorem{proposition}{Proposition}
\newcommand{\paperTitle}{GAAFormer: Geometry-Anchored Attention Transformer for Joint Point Cloud Denoising and Upsampling}
\newcommand{\paperTitleDisplay}{GAAFormer: Geometry-Anchored Attention Transformer\\for Joint Point Cloud Denoising and Upsampling}
\hypersetup{pdftitle={\paperTitle},pdfauthor={Haoxi Tian}}
\graphicspath{{figures/}}
\setlist[itemize]{leftmargin=*,topsep=3pt,itemsep=2pt,parsep=0pt}
\newcommand{\CleanCDGain}{3.35}
\newcommand{\CleanHDGain}{19.34}

\newcommand{\CoverageGain}{5.83}
\newcommand{\CNNCDGain}{3.47}
\newcommand{\CNNHDGain}{2.12}
\newcommand{\CNNPtwoFGain}{5.02}
\newcommand{\RawCDGain}{4.49}
\newcommand{\RawHDGain}{19.08}
\newcommand{\MortonCDChange}{1.35}
\newcommand{\MortonHDChange}{6.51}
\newcommand{\HighNoiseCDChange}{3.43}

\newcommand{\UpsampleCDChange}{0.33}
\newcommand{\UpsampleHDGain}{0.16}
\newcommand{\UpsamplePtwoFChange}{0.71}

\def\confName{CVPR}
\def\confYear{2027}
\title{\paperTitleDisplay}
\author{Haoxi Tian\\
School of Civil Engineering and Geomatics\\
Southwest Petroleum University}
\begin{document}
\raggedbottom
\maketitle
\begin{abstract}
Point cloud denoising and upsampling require samples that accurately recover
the underlying surface from noisy or sparse observations. A central challenge
in both tasks is preserving surface coverage while refining point locations.
We introduce \method\ (Geometry-Anchored Attention Transformer), a unified generative octree transformer that performs
both tasks with a shared network. Its decoder anchors attention neighborhoods
to geometry, providing spatial context for occupancy prediction and point
generation. Fixed voxel windows and alternating shifts preserve the grouping
of surviving nodes as the output tree changes. The design retains the encoder,
positional encoding, prediction heads, and parameter count of a
serialized-attention backbone. We establish a deletion-consistency property
of the neighborhood rule and evaluate both tasks using the same model.
On PUNet, \method\ achieves a mean Chamfer distance (CD) of
$6.0643\times10^{-5}$, reducing CD and Hausdorff distance by
\CleanCDGain\% and \CleanHDGain\%, respectively, relative to the serialized
backbone. For upsampling, it achieves CD values of $0.8475$, $1.7057$,
and $1.0147$ ($\times10^{-5}$) on PU-GAN, PU1K, and Sketchfab,
respectively, with improvements in CD, Hausdorff distance, and point-to-surface
distance over the same backbone on PU-GAN and Sketchfab.
These results establish a single architecture for point cloud denoising
and upsampling. Within the shared encoder--decoder, the same
geometry-anchored attention module improves both noise removal and dense
point generation without task-specific redesign or additional parameters.
\end{abstract}

\section{Introduction}
\label{sec:intro}
\begin{figure*}[t]
    \centering
    \includegraphics[width=\textwidth]{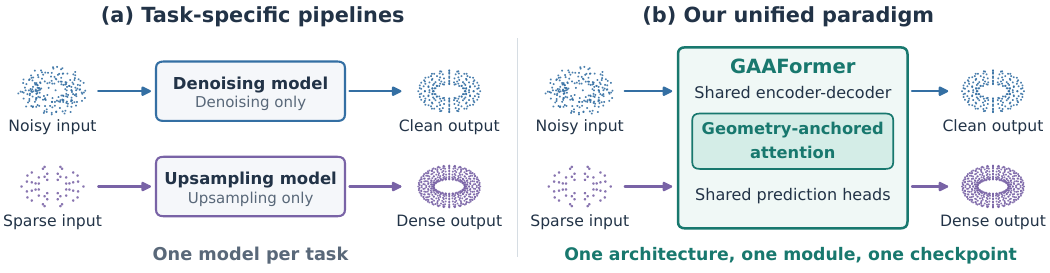}
    \caption{\textbf{Unified denoising and upsampling.}
    Task-specific models serve one task each (left).
    \method\ uses the same architecture, attention module, and checkpoint
    for both tasks (right). Point clouds are schematic.}
    \label{fig:paradigm}
\end{figure*}

Point cloud denoising removes geometric noise, while point cloud upsampling
generates dense samples from sparse observations. Both tasks require accurate
recovery of the underlying surface: predicted points must lie near the surface
and cover its geometry. A unified network must therefore suppress noise and
generate new samples while preserving geometric detail and surface coverage.
\Cref{fig:paradigm} contrasts task-specific pipelines with our unified
architecture, in which the same geometric attention module and network
weights support both tasks.

Generative octree decoders provide a shared representation for these tasks.
They predict occupied cells from coarse to fine and place a point in each
retained leaf, allowing the output support and cardinality to differ from
the input~\cite{ocnn,outputguided}. Transformer backbones add feature
interactions over groups of octree nodes. Serialized neighborhoods make
these interactions efficient~\cite{octformer,ptv3}, but their membership
depends on positions in a compact node sequence. In a generative decoder,
this sequence is rebuilt as predicted occupancy changes.

The central problem is preserving reliable geometric context while the
decoder changes its own set of active cells. Pruning a branch can shift
surviving nodes across serialized group boundaries even when their coordinates
are unchanged, altering the context used to predict occupancy and point
locations. An incorrect coarse occupancy decision removes all descendant
samples, so finer coordinate refinement cannot restore the excluded region.
This creates a shared difficulty: denoising must suppress perturbations while
retaining valid surface regions, whereas upsampling must extend geometric
evidence into sparsely sampled regions. Both tasks therefore require spatially
consistent attention neighborhoods throughout tree generation.

To address this problem, we introduce \method\ (Geometry-Anchored Attention Transformer), a unified generative octree
transformer whose decoder anchors attention groups to spatial coordinates.
Each cell is assigned to a fixed $4\times4\times4$ voxel window, and
consecutive blocks alternate the window shift to exchange information across
boundaries. At a fixed depth and shift, group assignment depends only on
coordinates, so removing nodes does not regroup the survivors.
One encoder--decoder, shared prediction heads, and a single checkpoint serve
both denoising and upsampling.

We evaluate \method\ on PUNet for denoising and on PU-GAN, PU1K, and
Sketchfab for upsampling. The study includes quantitative and visual
comparisons, ablations of attention layout, window capacity, training
objectives, and feature access, and measurements of neighborhood stability
and inference cost.

Our contributions are:
\begin{itemize}
    \item We introduce \method, a unified generative octree transformer
    for point cloud denoising and upsampling, with one encoder--decoder,
    shared prediction heads, and a single checkpoint.
    \item We propose a shared geometry-anchored attention module that improves
    PUNet denoising and PU-GAN and Sketchfab upsampling within the same
    architecture, without task-specific redesign or additional parameters.
    \item We establish deletion-consistent attention membership and validate
    the unified model through benchmark comparisons, architectural ablations,
    neighborhood perturbations, and inference-cost measurements.
\end{itemize}

\section{Related Work}
\label{sec:related}
\smallhead{Point cloud denoising and upsampling.}
Learning-based upsampling constructs dense samples from sparse observations.
PU-Net~\cite{punet} expands learned point features, PU-GAN~\cite{pugan}
introduces adversarial distribution learning, and PU-GCN~\cite{pugcn}
uses graph convolutions and NodeShuffle. MPU~\cite{mpu} and
PU-CRN~\cite{pucrn} refine points progressively, while
PU-Transformer~\cite{putransformer} explores transformer feature extraction.
Neural Points~\cite{neuralpoints} and iPUNet~\cite{ipunet} learn continuous
local surfaces; Grad-PU~\cite{gradpu} refines interpolated points using a
learned distance field. For denoising, PointCleanNet~\cite{pointcleannet} learns outlier removal and
point correction; differentiable manifold reconstruction~\cite{dmr}
reconstructs local surfaces before resampling; and score-based
denoising~\cite{score} iteratively updates points using an estimated score
field. PointFilter~\cite{pointfilter} predicts point displacements,
PCDNF~\cite{pcdnf} couples denoising with normal filtering, and
IterativePFN~\cite{iterativepfn} learns successive filtering stages.
Our focus is the representation of context inside a hierarchical
decoder that generates output support.

\smallhead{Octree representations and generation.}
O-CNN~\cite{ocnn} makes high-resolution 3D convolution practical by organizing
features in an octree. Output-guided skip connections~\cite{outputguided}
align encoder information to a generated output tree, enabling shape and
scene completion. The same output-guided formulation has been applied to
joint point cloud upsampling and cleaning~\cite{ounet}. We retain its tree construction,
scale transitions, aligned skip connections, occupancy heads, and leaf
coordinate prediction. Our contribution concerns the decoder's attention
neighborhoods, which must remain spatially meaningful as the generated
support changes.

\smallhead{Attention neighborhoods for 3D data.}
Point Transformer~\cite{pointtransformer} develops attention over local point
neighborhoods. Stratified Transformer~\cite{stratified} combines dense nearby
keys with sparse distant keys. OctFormer~\cite{octformer} partitions sorted
octree keys into fixed-cardinality groups, while Point Transformer
V3~\cite{ptv3} further demonstrates the scalability of serialization.
Spatial windows and shifts originate in Swin Transformer~\cite{swin};
Swin3D~\cite{swin3d} extends window-based processing to sparse 3D
representations. Thus, spatial windows themselves are an established
primitive. We study their particular role when the network generates and
prunes the domain on which its attention operates, and give an explicit
membership property for that changing domain.

\smallhead{Position encoding and neighborhood construction.}
Relative positional information can distinguish locations within an
attention group, while convolutional positional encoding introduces local
features~\cite{octformer,stratified}. These operations complement the choice
of which keys enter a group. In \method, the original positional encoding
and projection parameters remain in place; coordinate anchoring changes
the support of the attention operator. This separation enables direct
initialization from the serialized backbone and direct replacement of
its grouping rule.

\section{Method}
\label{sec:method}

\subsection{Motivating Observations}
\label{sec:observations}
In both denoising and upsampling, the generated points should recover the
geometry and coverage of the underlying surface. We examine how octree
generation and attention neighborhoods affect this shared requirement.

\smallhead{Surface loss from coarse pruning.}
Let a decoder expand occupied octree cells from depth $d_0$ to depth $D$.
If a cell at depth $d$ is rejected, none of its descendants can appear in
the generated tree. A local occupancy error therefore has a spatial extent
that depends on its depth. Coarse decisions determine the spatial support
available for subsequent refinement, while the finest coordinate head
positions samples inside the retained leaves. Geometric context should
therefore support occupancy prediction throughout the hierarchy.

\smallhead{Neighborhood sensitivity to tree updates.}
\Cref{fig:stability} summarizes a perturbation study of the grouping rule. We delete 1\%
of depth-five branches in six training geometries and inspect 4,096 queries
at each of depths six through eight. After removing deleted neighbors from
the comparison, the mean neighborhood Jaccard is 0.608 for regular Morton
groups and 0.546 for dilated groups. Coordinate windows preserve a Jaccard
of 1.000. These observations motivate spatially bounded attention
neighborhoods whose membership follows the coordinates of the generated
cells as the output support changes.

\begin{figure*}[t]
    \centering
    \includegraphics[width=0.9\textwidth]{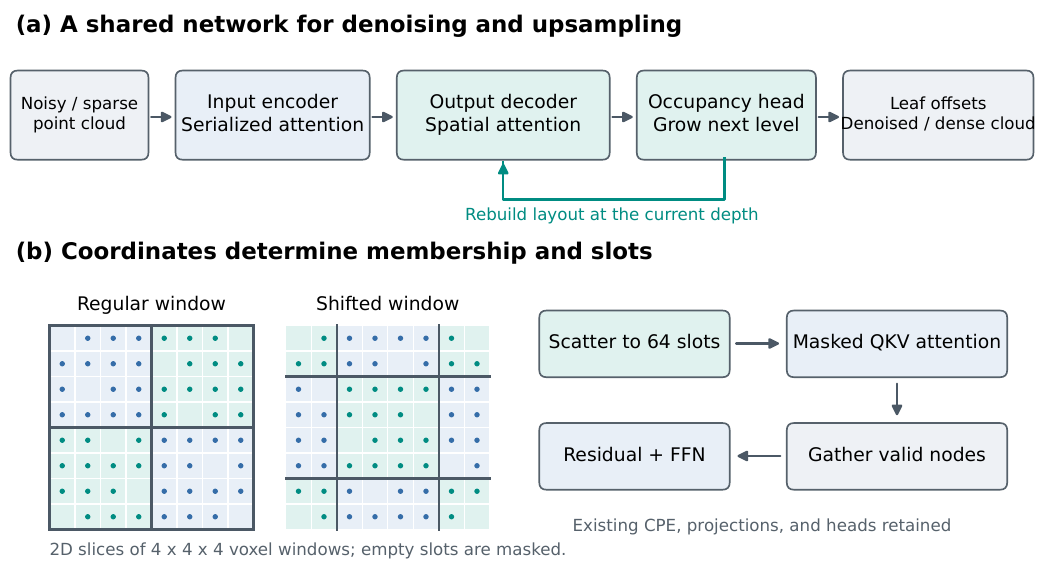}
    \caption{\textbf{\method\ for denoising and upsampling.}
    (a) A shared encoder processes the octree of a noisy or sparse input.
    The output-guided decoder predicts occupancy and grows the output tree;
    its attention layout is constructed from the nodes at the current depth.
    (b) Coordinate-based windows alternate with a two-voxel shift. The
    schematic shows 2D slices of the 3D windows. Scatter, masked attention,
    and gather retain the backbone's projections and residual processing.}
    \label{fig:architecture}
\end{figure*}

\subsection{Output-Guided Octree Backbone}
\label{sec:backbone}
Given a noisy or sparse point cloud $P=\{p_i\}_{i=1}^{N}$, we construct an input
octree $\T_{\mathrm{in}}$ with finest depth $D=8$ and full depth $d_0=2$.
The encoder produces features $E^d$ at depths $d=2,\ldots,8$. The output
octree starts from a full grid at depth two and grows according to predicted
occupancy. At a subsequent depth, decoder features are computed as
\begin{equation}
 H^d = B_d\!\left(
       U_d(H^{d-1}) +
       \operatorname{Align}(E^d,\T_{\mathrm{in}},\T_{\mathrm{out}})
       \right),
 \label{eq:decoder}
\end{equation}
where $U_d$ is octree upsampling, $\operatorname{Align}$ matches input
features to current output keys, and $B_d$ is a two-block transformer stage.
At $d_0$, the stage receives the corresponding encoder features directly.
We use established output-guided alignment and scale
transitions~\cite{outputguided,ounet}.

An occupancy head produces two logits $z_i^d$ for every candidate cell.
The cell is retained when $z_{i,1}^d>z_{i,0}^d$. Retained cells generate
eight child candidates until the finest depth. At depth $D$, a coordinate
head predicts an offset
\begin{equation}
 \delta_i=\tfrac12\tanh(h_{\mathrm{coord}}(H_i^D)),
 \qquad \delta_i\in(-\tfrac12,\tfrac12)^3.
\end{equation}
Adding this offset to the leaf center produces one output point per retained
leaf, followed by conversion to the evaluation coordinate system.
Consequently, output cardinality is determined by generated occupancy.
This shared generation process produces denoised samples from noisy inputs
and dense samples from sparse inputs.

The encoder uses regular and dilated octree attention with 32-token
serialized groups and dilation four. \method\ changes the decoder grouping.
The convolutional positional encoding (CPE), relative positional encoding
(RPE), query/key/value projections, feed-forward networks, and prediction
heads retain their existing parameters. CPE is a depthwise $3^3$ octree
convolution followed by octree GroupNorm. Each transformer block applies
the CPE residual, pre-normalized attention, and a pre-normalized pointwise
MLP residual. This combines convolutional local features with attention
over explicitly defined spatial groups.

\subsection{Geometry-Anchored Window Attention}
\label{sec:attention}
At depth $d$, let $\U_d$ denote the current candidate cells. A cell $i$
has integer coordinates $x_i\in\{0,\ldots,2^d-1\}^3$ and batch index $b_i$.
For window width $w=4$ and shift $s\in\{0,2\}$, define
\begin{align}
 g_s(x_i)&=\left\lfloor\frac{x_i+s\mathbf{1}}{w}\right\rfloor,
 \label{eq:group}\\
 \ell_s(x_i)&=(x_i+s\mathbf{1})\bmod w.
 \label{eq:slot}
\end{align}
The group identifier is $(b_i,g_s(x_i))$. The local coordinate assigns the
node to slot $16\ell_x+4\ell_y+\ell_z$ in a 64-slot buffer. An octree cell
has a unique coordinate at a given depth, so valid nodes never collide
within a slot. Grouping also keeps different objects separate.

We gather cells with the same group identifier and mask unoccupied buffer
slots. For one attention head with feature dimension $h$, the output is
\begin{align}
 a_{ij} &=
 \frac{\exp(q_i^\top k_j/\sqrt{h}+r(x_i-x_j))}
 {\sum_{k\in\N_s(i;\U_d)}
 \exp(q_i^\top k_k/\sqrt{h}+r(x_i-x_k))}, \label{eq:weights}\\
 y_i &= \sum_{j\in\N_s(i;\U_d)} a_{ij}v_j, \label{eq:attn}\\
 \N_s(i;\U_d)&=\{j\in\U_d:b_j=b_i,\ g_s(x_j)=g_s(x_i)\}.
 \label{eq:neighborhood}
\end{align}
Here $q,k,v$ are learned linear projections and $r$ is the inherited
relative positional bias. The implementation masks invalid keys before
softmax and discards padded query outputs during gathering. Since every
valid query occupies its own group, its neighborhood is nonempty.

The first block uses $s=0$ and the second uses $s=2$.
At depth two, both use $s=0$, allowing all 64 cells of the full grid to
interact. Shifts translate the partition boundaries; they do not wrap the
spatial domain. Alternating partitions allow information to cross regular
window boundaries without a global attention matrix. We rebuild the
layout at the current decoder depth after tree growth, so masks and slot
indices always correspond to the available candidate cells.

\smallhead{A physical bound on context.}
Coordinate anchoring also bounds the spatial extent of every attention
edge. Two cells in one window differ by at most three voxels along each
axis. If $h_d=2^{1-d}/1.7$ is the cell width in evaluation coordinates,
their centers satisfy
\begin{equation}
 \|p_i^{\mathrm{center}}-p_j^{\mathrm{center}}\|_2
 \leq 3\sqrt{3}\,h_d .
 \label{eq:spatialbound}
\end{equation}
The bound holds for both shifts and every sparsity pattern. Serialized
groups constrain token count instead, so their physical extent varies
with the occupied support. The two shifted blocks expand communication
beyond a single window while retaining a geometric interpretation for
each interaction. This is an attention-edge bound, distinct from the
receptive field of the surrounding convolutional and residual operators.

\subsection{Deletion Consistency and Computational Cost}
\label{sec:property}
\begin{proposition}[Deletion-consistent membership]
Fix an octree depth, a shift, and the coordinates and batch indices of
the nodes. For any removed set $R\subseteq\U_d$ and surviving query
$i\in\U_d\setminus R$,
\begin{equation}
 \N_s(i;\U_d\setminus R)=\N_s(i;\U_d)\setminus R.
 \label{eq:consistency}
\end{equation}
\end{proposition}
\begin{proof}
Membership in \cref{eq:neighborhood} depends only on the query and key
coordinates and their batch indices. Removing other nodes changes none of
these quantities. The remaining members are therefore exactly the original
members outside $R$.
\end{proof}

This property rules out regrouping caused solely by compaction of the active
node list. It concerns the attention graph at one layer; attention weights
can still change when a key in the same window is deleted, and the full
network can change features through its other operators. For comparison,
a fixed-cardinality serialized partition can move a surviving node across
a group boundary after an earlier entry is removed. The proposition
formalizes the membership distinction without imposing assumptions on
learned feature values.

For $G_d$ nonempty windows and channel width $C$, the attention score and
value operations cost $O(G_d\,64^2 C)$, and the padded feature buffer has
$64G_d$ entries. Since the window capacity is fixed and $G_d\leq|\U_d|$,
this component scales linearly in the number of cells, with a
padding-dependent constant. Coordinate grouping is an additional
implementation cost; the expression does not assert a linear-time sorting
routine. The new grouping adds no learned tensors. Both backbones have
10,623,997 parameters, although their arithmetic and memory access differ.
We measure these differences in \cref{sec:cost}.

\begin{figure*}[t]
    \centering
    \includegraphics[width=0.86\textwidth]{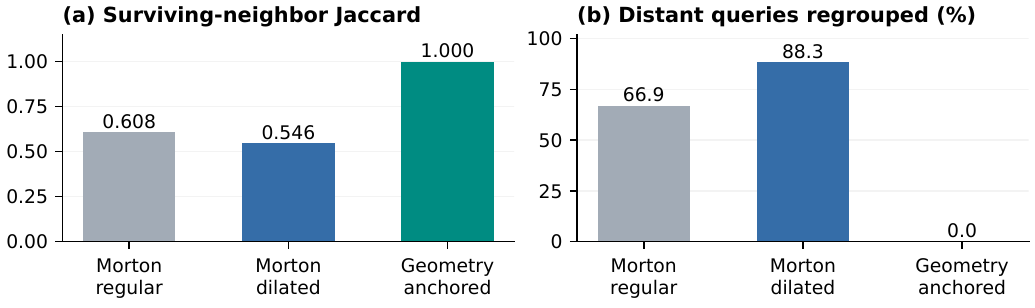}
    \caption{\textbf{Attention membership under branch removal.}
    After deleting 1\% of depth-five branches, we compare neighborhoods of
    surviving queries at depths six through eight. Deleted keys are removed
    from the reference neighborhood. Coordinate windows preserve the
    remaining membership, while serialized partitions regroup queries.}
    \label{fig:stability}
\end{figure*}

\subsection{Training Objective}
\label{sec:objective}
Training uses the ground-truth output octree to provide candidate cells.
Let $y_i^d$ be its occupancy label, $V_d$ its candidate set at depth $d$,
and $V_D^+$ its occupied finest cells. The original objective is
\begin{equation}
 \mathcal{L}
 =\sum_{d=d_0}^{D}\frac{1}{|V_d|}
   \sum_{i\in V_d}\operatorname{CE}(z_i^d,y_i^d)
 +\frac{1}{|V_D^+|}\sum_{i\in V_D^+}
   \|\delta_i-\delta_i^\star\|_2^2,
 \label{eq:loss}
\end{equation}
where $\delta_i^\star$ is the target leaf offset. At inference, the decoder
uses its own occupancy predictions and zero logit bias. The primary model
requires one backbone and no geometric output from a reference network.
Loss modifications are evaluated separately in \cref{sec:ablation}.

\section{Experiments}
\label{sec:experiments}

\begin{table*}[t]
    \centering
    \caption{\textbf{Point cloud denoising.}
    CD $\times10^5$ and P2F $\times10^3$.}
    \label{tab:denoising}
    \footnotesize
    \setlength{\tabcolsep}{3pt}
    \begin{tabular}{@{}lrrrrrrrrrrrr@{}}
\toprule
Method & \multicolumn{6}{c}{10k input points} & \multicolumn{6}{c}{50k input points} \\
\cmidrule(lr){2-7}\cmidrule(lr){8-13}
& \multicolumn{2}{c}{1\% noise} & \multicolumn{2}{c}{2\% noise} & \multicolumn{2}{c}{2.5\% noise}
& \multicolumn{2}{c}{1\% noise} & \multicolumn{2}{c}{2\% noise} & \multicolumn{2}{c}{2.5\% noise} \\
\cmidrule(lr){2-3}\cmidrule(lr){4-5}\cmidrule(lr){6-7}\cmidrule(lr){8-9}\cmidrule(lr){10-11}\cmidrule(lr){12-13}
& CD & P2F & CD & P2F & CD & P2F & CD & P2F & CD & P2F & CD & P2F \\
\midrule
PointCleanNet~\cite{pointcleannet} & 11.22 & 2.45 & 14.71 & 3.81 & 16.75 & 4.87 & 2.81 & 1.54 & 4.13 & 3.35 & 5.17 & 4.93 \\
ScoreDenoise~\cite{score} & 9.32 & 2.09 & 13.40 & 3.60 & 16.44 & 3.86 & 2.58 & 1.49 & 3.76 & 3.19 & 4.89 & 3.04 \\
PointFilter~\cite{pointfilter} & 9.30 & 1.85 & 13.82 & 2.90 & 15.35 & 3.43 & 2.74 & 1.10 & 3.90 & 1.82 & 4.18 & 2.11 \\
PCDNF~\cite{pcdnf} & 9.12 & 1.72 & 13.72 & 2.75 & 15.04 & 3.17 & 2.52 & 1.00 & 3.48 & 1.86 & 3.89 & 2.31 \\
IterativePFN~\cite{iterativepfn} & 7.75 & 1.29 & 12.21 & 2.13 & 13.48 & 2.58 & 2.42 & 0.82 & 3.19 & 1.50 & 3.78 & 2.11 \\
OUNet~\cite{ounet} & 7.59 & 1.52 & 9.62 & 2.49 & 10.30 & 2.95 & 1.78 & 0.94 & 2.61 & 1.51 & 3.13 & 1.95 \\
\midrule
Serial-OUFormer & 7.7954 & 1.5433 & 9.6666 & 2.5594 & 10.9477 & 2.9540 & 1.7070 & 0.8858 & 2.8460 & 1.5869 & 4.6830 & 1.9321 \\
\textbf{GAAFormer} & 7.6962 & 1.5286 & 9.6720 & 2.5747 & 11.3226 & 2.9556 & 1.7087 & 0.8887 & 2.6545 & 1.5893 & 3.3320 & 1.9099 \\
\bottomrule
\end{tabular}

\end{table*}

\begin{table*}[t]
    \centering
    \caption{\textbf{Point cloud upsampling.}
    CD $\times10^5$, HD $\times10^3$, and P2F $\times10^3$.}
    \label{tab:upsampling}
    \small
    \setlength{\tabcolsep}{4.5pt}
    \begin{tabular}{@{}lrrrrrrrrr@{}}
\toprule
Method & \multicolumn{3}{c}{PU-GAN (27 inputs)} & \multicolumn{3}{c}{PU1K (100 inputs)} & \multicolumn{3}{c}{Sketchfab (14 inputs)} \\
\cmidrule(lr){2-4}\cmidrule(lr){5-7}\cmidrule(lr){8-10}
& CD & HD & P2F & CD & HD & P2F & CD & HD & P2F \\
\midrule
PU-Net~\cite{punet} & 3.095 & 2.0468 & 2.875 & 3.622 & 2.5043 & 4.717 & 4.194 & 2.1651 & 3.682 \\
iPUNet~\cite{ipunet} & 2.215 & 1.4163 & 2.247 & -- & -- & -- & -- & -- & -- \\
MPU~\cite{mpu} & 1.492 & 1.4574 & 1.361 & 2.860 & 2.0772 & 4.098 & 2.320 & 1.5937 & 2.085 \\
PU-GCN~\cite{pugcn} & 1.454 & 1.4438 & 1.323 & 2.675 & 1.8529 & 3.975 & 2.364 & 1.7232 & 2.118 \\
PU-GAN~\cite{pugan} & 1.318 & 1.1366 & 1.230 & 2.443 & 1.7182 & 3.811 & 2.005 & 1.1744 & 1.893 \\
PU-CRN~\cite{pucrn} & 0.845 & 0.5741 & 0.549 & 2.241 & 0.9659 & 3.538 & 1.261 & 0.8717 & 1.050 \\
Neural Points~\cite{neuralpoints} & 1.100 & 1.6809 & 0.778 & -- & -- & -- & 0.943 & 1.0123 & 0.616 \\
Grad-PU~\cite{gradpu} & 1.150 & 0.2318 & 0.690 & 3.997 & 0.4627 & 3.779 & 1.346 & 0.2440 & 1.048 \\
OUNet~\cite{ounet} & 0.801 & 0.1044 & 0.425 & 1.818 & 0.3306 & 3.458 & 1.004 & 0.1869 & 0.768 \\
\midrule
Serial-OUFormer & 0.8479 & 0.1021 & 0.3327 & 1.6987 & 0.3085 & 0.6689 & 1.0154 & 0.2011 & 0.6588 \\
\textbf{GAAFormer} & 0.8475 & 0.0987 & 0.3320 & 1.7057 & 0.3099 & 0.6754 & 1.0147 & 0.1929 & 0.6569 \\
\bottomrule
\end{tabular}

\end{table*}

\subsection{Experimental Setup}
\smallhead{Tasks and data.}
We evaluate denoising on 20 PUNet test geometries under six conditions: 10k or 50k
input points with 1\%, 2\%, or 2.5\% noise, giving 120 inputs. Upsampling
uses 27 PU-GAN, 100 PU1K, and 14 Sketchfab inputs, giving 141 inputs.
The same \method\ checkpoint handles all 261 denoising and upsampling inputs.

\smallhead{Implementation.}
We optimize the models for 1,024 steps using AdamW with learning rate
$10^{-5}$, weight decay 0.05, and batch size two, and evaluate the final
checkpoints.

\smallhead{Metrics.}
CD is the sum of the two mean squared nearest-neighbor distances.
HD is the maximum squared nearest-neighbor distance over both directions.
P2F is mean Euclidean distance to the reference mesh. We report
CD $\times10^5$, HD $\times10^3$, and P2F $\times10^3$; lower is better.
Farthest-point sampling (FPS) caps predictions at 10k or 50k points for
denoising and 80k for upsampling. Predictions already below the cap are
unchanged. Raw evaluations retain every generated point.

\smallhead{Comparison methods.}
We compare with representative denoising and upsampling methods, including
OUNet~\cite{ounet} as a representative octree-convolutional network.
External comparison values in \cref{tab:denoising,tab:upsampling} are taken
from its reported benchmark.
We denote the counterpart with serialized decoder attention by \serial.
\method\ replaces its decoder grouping while retaining the encoder,
positional encodings, and prediction heads.
Missing entries are shown as dashes; HD values are converted to the
display units above.

\subsection{Denoising and Upsampling Results}
\label{sec:restoration}
\smallhead{Denoising.}
\Cref{tab:denoising} compares \method\ with PointCleanNet, ScoreDenoise,
PointFilter, PCDNF, IterativePFN, and OUNet.
Its CD is lower than the reported values of the first five methods in all
six conditions. IterativePFN has lower P2F in five conditions.
Compared with reported OUNet, \method\ achieves lower CD at 50k points
and 1\% noise.

Across the 120 denoising inputs, \method\ achieves CD 6.0643,
HD 0.6655, and P2F 1.9078. Relative to \serial, CD decreases
from 6.2743 and HD from 0.8251, giving improvements of
\CleanCDGain\% and \CleanHDGain\%, respectively. Mean P2F changes
from 1.9102 to 1.9078. The reference-to-prediction component of CD
improves by \CoverageGain\%, showing more accurate surface coverage.
Relative to our OUNet checkpoint, the reductions are \CNNCDGain\% CD,
\CNNHDGain\% HD, and \CNNPtwoFGain\% P2F.

\begin{figure*}[t]
    \centering
    \includegraphics[width=\textwidth]{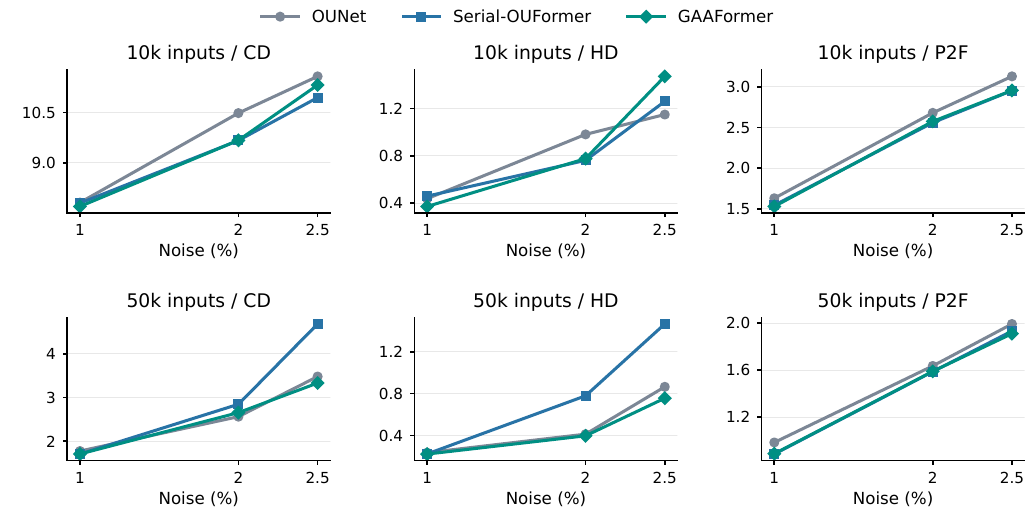}
    \caption{\textbf{Denoising across noise levels and input densities.}
    Each point averages 20 geometries. The top and bottom rows use 10k
    and 50k input points, respectively. CD, HD, and P2F are scaled by
    $10^5$, $10^3$, and $10^3$.}
    \label{fig:denoisingtrends}
\end{figure*}

\Cref{fig:denoisingtrends} summarizes performance across input densities
and noise levels. The largest CD and HD gains over \serial\ occur
at 50k points with 2.5\% noise. At 10k points and 2.5\% noise,
CD increases by \HighNoiseCDChange\%; the 10k, 2\%-noise setting
also slightly favors the serialized backbone.
Over all denoising inputs, \method\ improves CD on 67, HD on 66,
and P2F on 68. On raw predictions, aggregate CD and HD decrease by
\RawCDGain\% and \RawHDGain\%, respectively.

\smallhead{Upsampling.}
\Cref{tab:upsampling} includes feature expansion, progressive refinement,
neural-surface, and gradient-based approaches. \method's CD/HD/P2F
on PU-GAN are 0.8475/0.0987/0.3320, compared with the reported
0.845/0.5741/0.549 for PU-CRN and 1.150/0.2318/0.690 for Grad-PU.
On Sketchfab, Neural Points reports lower CD and P2F; on PU1K, all three
of \method's values are below the reported entries.

Relative to \serial, \method\ improves all three metrics on PU-GAN
and Sketchfab. On PU1K, CD, HD, and P2F increase by 0.41\%, 0.48\%,
and 0.97\%, respectively. Weighted over all 141 inputs, CD and P2F
increase by \UpsampleCDChange\% and \UpsamplePtwoFChange\%, while
HD improves by \UpsampleHDGain\%. The decoder thus preserves broadly
similar upsampling accuracy alongside its denoising improvements.

\subsection{Visual Comparisons}
\label{sec:qualitative}
\begin{figure*}[!t]
    \centering
    \includegraphics[width=0.96\textwidth]{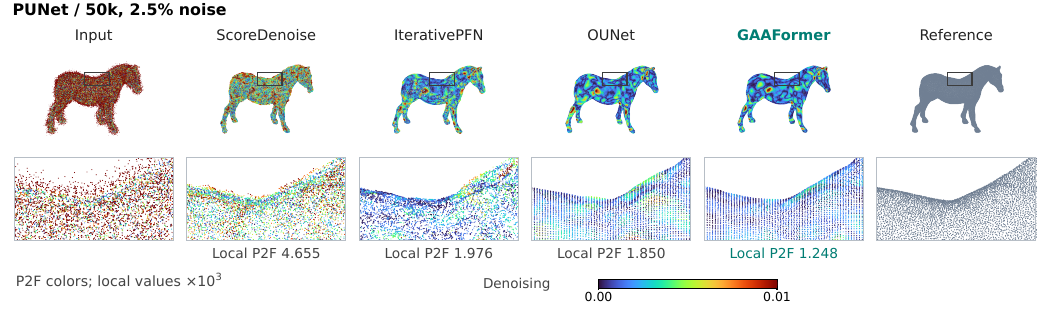}
    \caption{\textbf{Visual comparison for point cloud denoising.}
    PUNet with 50k input points and 2.5\% noise. Rectangles mark enlarged
    regions. Colors show point-to-surface distance; references are gray.
    Local P2F ($\times10^3$) averages predicted-point errors within each
    enlarged region.}
    \label{fig:denoisingvisual}
\end{figure*}

\begin{figure*}[!t]
    \centering
    \includegraphics[width=0.96\textwidth]{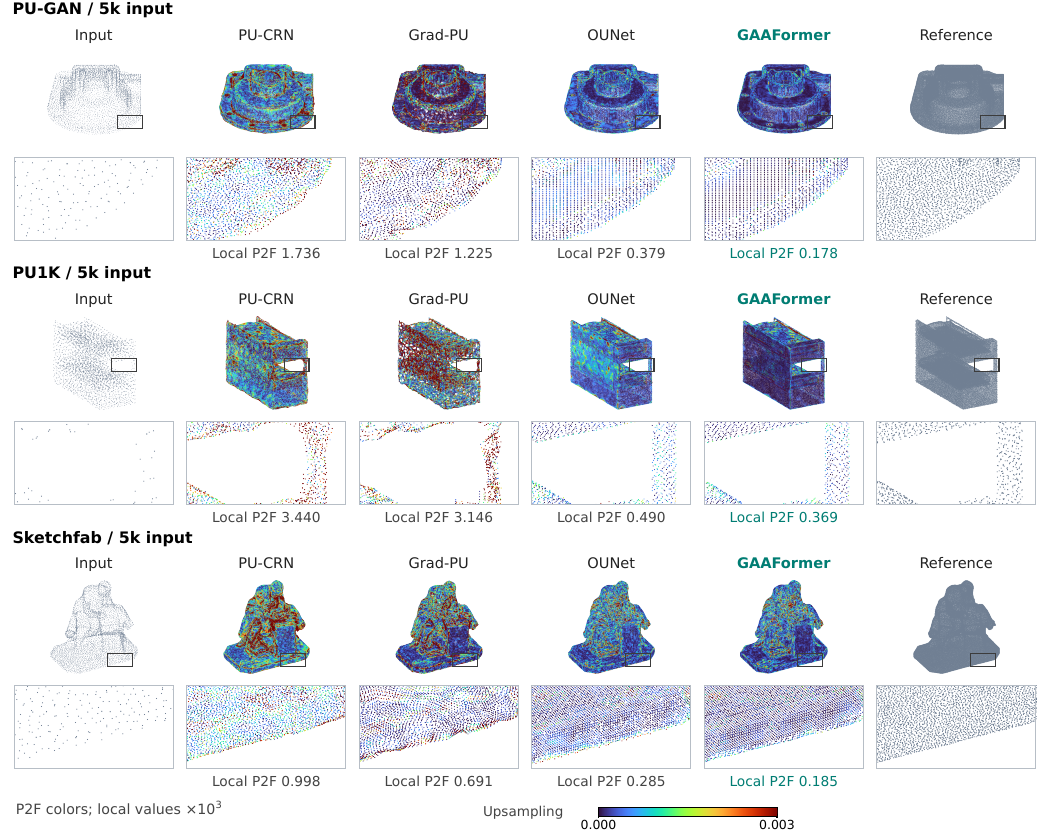}
    \caption{\textbf{Visual comparison for point cloud upsampling.}
    The three rows use 5k input points from PU-GAN, PU1K, and Sketchfab.
    Rectangles mark enlarged regions. Colors show point-to-surface distance;
    inputs and references are gray. Local P2F ($\times10^3$) averages
    predicted-point errors within each enlarged region.}
    \label{fig:upsamplingvisual}
\end{figure*}

\Cref{fig:denoisingvisual,fig:upsamplingvisual} compare selected challenging examples of
PUNet denoising and upsampling on PU-GAN, PU1K, and Sketchfab using methods from
\cref{tab:denoising,tab:upsampling}. ScoreDenoise and IterativePFN
denoise; PU-CRN and Grad-PU upsample; OUNet supports both tasks.
The enlarged regions expose residual scatter and sample displacement
along structural boundaries. In these regions, \method\ concentrates
predictions closer to the reference geometry and achieves lower local
P2F than the displayed baselines. Colors show point-to-surface distance;
the inset values quantify the corresponding local error.

\subsection{Ablation Studies}
\label{sec:ablation}
\begin{table}[t]
    \centering
    \caption{\textbf{Architecture and objective ablations.}
    Mean denoising CD $\times10^5$, HD $\times10^3$, and P2F $\times10^3$.}
    \label{tab:ablation}
    \small
    \setlength{\tabcolsep}{4pt}
    \begin{tabular}{lrrr}
\toprule
Method & CD & HD & P2F\\
\midrule
OUNet & 6.2826 & 0.6799 & 2.0085 \\
Serial-OUFormer & 6.2743 & 0.8251 & 1.9102 \\
Morton64 & 6.3589 & 0.8788 & 1.9088 \\
Serial + path & 6.2619 & 0.8216 & 1.9102 \\
Serial + coverage & 6.3014 & 0.8460 & 1.9021 \\
Serial + fine evidence & 6.2791 & 0.8299 & 1.9112 \\
\textbf{GAAFormer} & 6.0643 & 0.6655 & 1.9078 \\
GAA + path & 6.0534 & 0.6542 & 1.9079 \\
\bottomrule
\end{tabular}

\end{table}

\smallhead{Attention layout and window capacity.}
Morton64 increases serialized group capacity from 32 to 64 while retaining
the original grouping rule and dilation. Its mean CD and HD increase
by \MortonCDChange\% and \MortonHDChange\% relative to \serial\
(\cref{tab:ablation}). \method\ instead achieves CD 6.0643 and
HD 0.6655 with coordinate-based windows. These results show that
spatial anchoring provides benefits beyond enlarging the attention buffer.

\smallhead{Training objectives and feature access.}
The path variant adds a soft Dice loss on products of ancestor occupancy
probabilities. The coverage variant adds a differentiable margin-based
coverage objective. The fine-evidence variant introduces attention to
child-level encoder features before selected occupancy decisions.
Their denoising CD values range from 6.2619 to 6.3014.
Combining spatial attention with the path loss yields CD 6.0534 and
HD 0.6542, further improving on \method's original objective.

\subsection{Attention Neighborhood Stability}
\label{sec:stability}

\Cref{fig:stability} aggregates the layout study over six geometries.
Coordinate windows retain a Jaccard of 1.000 for both shifts.
Regular and dilated Morton groups obtain 0.608 and 0.546.
Among queries whose depth-five ancestor is more than one cell away
in Chebyshev distance from every removed branch, 66.9\% and 88.3\%
change group membership, respectively; the corresponding rate for
coordinate windows is zero. These results demonstrate the stability
of spatial grouping as the output support changes and agree with
the membership property in \cref{eq:consistency}.

\subsection{Inference Cost}
\label{sec:cost}

We measure inference on an RTX 4090 for a 10k
denoising input, a 50k denoising input, and a PU-GAN upsampling input.
Latency is the median of three synchronized repeats after two warm-ups.
Relative to \serial, \method\ changes latency from 131.02 to 135.79\,ms,
from 251.40 to 259.94\,ms, and from 113.68 to 124.82\,ms, respectively.
Peak allocated memory decreases from 1,273.7 to 1,043.3\,MiB,
from 2,341.8 to 2,057.8\,MiB, and from 1,342.9 to 1,091.9\,MiB.
The two architectures each contain 10,623,997 parameters.

\section{Conclusion}
\label{sec:conclusion}
We presented \method, a unified generative octree transformer for point
cloud denoising and upsampling. Its geometry-anchored attention module
combines fixed windows and alternating shifts to preserve surviving-cell
neighborhood membership and exchange context across boundaries. The module
serves both tasks within one encoder--decoder with shared prediction heads
and a single checkpoint, without additional learned parameters.
On PUNet, the module reduces mean denoising CD and HD by \CleanCDGain\%
and \CleanHDGain\% over the serialized backbone. The same module improves
upsampling CD, HD, and P2F on PU-GAN and Sketchfab. Ablations identify
spatial grouping as the central source of denoising gains, and inference
measurements show lower peak memory. Shared attention improves surface
recovery in both tasks.

\onecolumn
\twocolumn
{\small
\bibliographystyle{ieeenat_fullname}
\bibliography{references}
}
\end{document}